\documentclass[11pt,a4paper]{article}
\usepackage[hyperref]{acl2020}
\usepackage{times}
\usepackage{latexsym}

\usepackage{graphicx}

\usepackage{microtype}

\aclfinalcopy 

\usepackage{amsmath}

\title{Reducing Non-Normative Text Generation from Language Models}

\author{Xiangyu Peng\Thanks{\ Equal contributions}   ,  Siyan Li\footnotemark[1]  ,  Spencer Frazier, and Mark Riedl\\
Georgia Institute of Technology\\
Atlanta, GA 30332\\
\texttt{\{xpeng62, sli613, sfrazier7, riedl\}@gatech.edu}}

\date{}

\definecolor{mygreen}{RGB}{50, 128, 50}

\newcommand{\GG}{\textit{Goofus \& Gallant}}
\newcommand{\GnG}{\textit{G\&G}}

\begin{document}
\maketitle
\begin{abstract}
Large-scale, transformer-based language models such as GPT-2 are pretrained on diverse corpora scraped from the internet. Consequently, they are prone to generating non-normative text (i.e. in violation of social norms). 
We introduce a technique for fine-tuning GPT-2, using a policy gradient reinforcement learning technique and a normative text classifier to produce reward and punishment values.
We evaluate our technique on five data sets using automated and human participant experiments. 
The normative text classifier is 81-90\% accurate when compared to gold-standard human judgements of normative and non-normative generated text.
Our normative fine-tuning technique is able to reduce non-normative text by 27-61\%, depending on the data set.

\end{abstract}

\section{Introduction}


Human societies implicitly establish codes of acceptable behavior in social contexts.
{\em Normativity} is behavior that conforms to expected societal norms and contracts, whereas non-normative behavior aligns to values that deviate from these expected norms.
\citet{sumner1967normative} defines norms as: ``...informal rules that are not written, but, when violated, result in severe punishments and social sanction upon the individuals, such as social and religious exclusions.'' 
Non-normativity does not connote behavior devoid of value or immoral, but behavior that fails to conform to social standards shared by other individuals in the relevant group, organization or society.
Norms can also be thought of as actions taken by an entity which conform to an identity \citep{katzenstein1996culture}, thus allowing others to categorize behavior as in-group or out-group.
Different societies and groups collectively have different ideals about what actions constitute normative behavior; group members use these ideals to heuristically guide their actions to avoid social ostracization.
For example, many societies have norms against violence, or certain behaviors being conducted in public.
Conflicts between individuals can arise when enacting non-normative behaviors or uttering non-normative speech.

This paper examines generative language models and the frequency at which they generate descriptions of non-normative behavior. 
Large-scale, transformer-based neural language models such as ELMo \cite{elmo}, BERT \cite{BERT}, GPT-2 \cite{radford2019language}, GPT-3, Grover \cite{Grover}, CTRL \cite{keskar2019ctrl}, T5 \cite{raffel2019exploring}, and XLNet \cite{yang2019xlnet} are trained on very large corpora such as text scraped from the internet, books, or both.
%

These language models generate text that is statistically representative of the corpora they were trained on.
As such, text scraped from the internet co-mingles text produced by many groups with differing norms, as well as text produced by people intentionally using non-normative speech, like ``trolling'' language. Models trained on these data can then produce undesirable, harmful output. 
Stories from the internet and books also contain normative and non-normative situations (e.g., antagonists, as well as protagonists conducting conventionally non-normative behaviors). 
Consequently, it is possible, and often likely, for language models to generate non-normative descriptions of behavior (murder, crime, suicide, racist actions, rude behavior, etc.), exhibit biases against certain demographics groups \cite{sheng2019woman, solaiman2019release}, stereotypical biases \cite{nadeem2020stereoset} or racist text when prompted with trigger phrases \cite{wallace2019universal}.
%
%

{\em Value alignment} \cite{russell2015research} is the concept that an agent is unable to perform actions that cause harm to humans. 
Harmful behavior is not limited to physical actions by robots, the focus of some AI value alignment research. We recognize that natural language communication can also cause harm.
For example, Amazon Alexa, a virtual assistant AI, was reported to suggest a user commit suicide.\footnote{\url{https://www.newsweek.com/amazon-echo\%2Dtells-uk-woman-stab-herself-1479074/}}
\citet{frazier2019learning} developed a classifier for normative behavior
which exhibits strong zero-shot and few-shot transfer across a variety of text corpora.
The authors speculate that their model---which they call a {\em value-aligned prior}---can bias model output perceived as more normative. 
In this paper we ask a different question: whether a value-aligned prior can be used to reduce the generation of descriptions of non-normative behavior by neural language models.

The common approach to fine-tuning language models is to provide additional corpora of exemplars.
If a corpus of exemplars is normative, the language model can be trained to emulate this over time. 
Generally, in the absence of very large normative corpora, we need an alternative approach to fine-tuning language models.
We use a reinforcement learning approach to fine-tuning language models, using the normative behavior classifier of~\citet{frazier2019learning} as a non-differentiable reward function. Our method back-propagates reward 
relative to the degree of non-normativity of text generated by the language model.

We evaluate our reinforcement learning fine-tuning technique with three sets of experiments.
First, we replicate the experiments by \citet{frazier2019learning} on text generated by a language model instead of originally held-out corpus text.
Second, we show with automated and human participant experiments that fine-tuning on reward generated by a normative classifier model can reduce the generation of non-normative text by $ 27-61\%$. 
Third, we ablate our technique and show with automated and human participant experiments that the fine-tuning technique works with classifiers other than the normative classifier---specifically models trained to classify negative-sentiment and toxic language. 

%
%



\section{Background and Related Work}

\subsection{Value Alignment and Normative Priors}

Humans have expectations that --- just like other humans --- agents will avoid harmful actions, conform to personal values and to social norms \cite{bicchieri2005grammar}, even when not explicitly communicated. This is referred to as the {\em value alignment  problem} \cite{soares2014aligning,russell2015research,arnold2017value,abel2016reinforcement}. 
Harmful agent behavior can theoretically be mitigated by casting values as preferences over action sequences. For example \citet{christiano2017deep} collected human preferences to shape rewards for game-playing agents in reinforcement learning. 

Instead of preference learning, \citet{frazier2019learning} used the BERT~\cite{BERT} 
language model's token embeddings to train a binary classifier. This  model is used to differentiate between normative and non-normative natural language sentences containing events, utterances and descriptions of behavior.
They obtained training data from \GG{} \textit{(\GnG)}, a children's educational comic strip featuring two characters of the same names.
Goofus always deviates from the ``proper" way to behave, while Gallant always performs the behavior of an exemplary child in western society at the time the comics were created. 
As a result, \GnG{} is a naturally labeled source of normative and non-normative text, for the specific society it represents. 


\citet{frazier2019learning} demonstrated this method could accurately classify descriptions of behavior as normative or non-normative.
Furthermore, this classifier retained high performance in zero-shot and few-shot transfer tasks. 
For example, they show that their classifier, trained on \GnG{} comics, can classify normative event descriptions in contemporary collections of popular plot points and science fiction plot summaries, instances of medium- and far-transfer, respectively. 
The authors speculate that their classifier model can bias agent behavior toward normative courses of action in other contexts.
However, this was not directly shown.  
We ask whether a normative classifier can be used to fine-tune the ``behavior'' of a large-scale transformer-based language model.

\subsection{Language Model Training \& Fine-Tuning}

Large-scale transformer-based neural language models such as BERT and GPT-2 are trained on large corpora of text scraped from the web and books. 
They can be fine-tuned to a specific domain of interest, 
commonly accomplished by providing a corpus of exemplars from that domain.
Over time, the weights of the pre-trained model will shift and increasingly generate passages which better emulate the corpus of exemplars.
If the fine-tuning corpus of exemplars is normative, the language model will, in theory, learn to prefer normative language over time.
\GG{} is one such normative corpus, and if a language model is fine-tuned on it then it may prefer to generate normative language.

GPT-2~\cite{radford2019language}, in particular, is a large-scale transformer-based language model trained on a large corpus of text scraped from web pages and social media. 
Applying the concept of value alignment as preference learning, 
\citet{ziegler2019fine} use a reinforcement learning method on the 774M-parameter version of GPT-2 to favor human-preferred text. 
Crowd workers were asked to select generated text completions from a set of given prompts that had positive sentiment.
These preference values were used to fine-tune GPT-2. 
This is one possible technique for reinforcement-based fine-tuning; sentiment is, however, not necessarily a good measure of adherence to norms.
We replace the linear reward model for fine-tuning GPT-2 \cite{ziegler2019fine} with a pre-trained normative text classifier.

The Plug and Play Language Models (PPLM) \cite{dathathri2019plug} also apply attribute classifiers to fine-tune language models; 
the technique is demonstrated via generating text with a target sentiment and also decreasing the frequency of toxic language.
There are two limitations: 
(a)~PPLM trains a model to operate on a {\em fixed} set of prefix input, and
(b)~the classification must be done on a word-by-word basis and thus cannot easily be applied to problems where
the normative valence of individual words relies on a single or multiple sentence context (e.g. quoting and admonishing toxic speech).
Our fine-tuning technique, in contrast, works on arbitrary prefixes and assesses the the normativity of entire sentences.



\subsection{Datasets}
\label{sec:Datasets}

We make use of five datasets, chosen to represent a diverse set of domains.
The normative text classifier by \citet{frazier2019learning} was tested on a corpus of science fiction plot summaries~\cite{ammanabrolu2019story} 
as well as a new {\em Plotto} dataset, based on a book by the same name that catalogues plot points for scaffolding fictional story-writing.
Story corpora are particularly good for testing problems pertaining to textual descriptions of normative and non-normative behavior. 
Stories contain antagonists that frequently violate societal norms and protagonists who are more likely to exemplify contemporary social norms. 
We recognize many stories require protagonists to perform non-normative behaviors like violence against others to achieve normative ends, further indicating the importance of accounting for a broader frame of context when determining normativity.

The science-fiction plot summary corpus~\cite{ammanabrolu2019story} is a collection of 2,276 stories scraped from crowd-sourced plot summaries on fan sites. 
These stories have an average length of 89.23 sentences. 
Sentences in this corpus tend to give high-level overviews of the actions that characters are performing (e.g.  ``Lyta accuses Sinclair of attempting to murder the ambassador").
The sci-fi corpus also presents a transfer challenge because it involves a lot of novel entities---aliens, spaceships, laser weapons, etc.---that do not exist in \GG.
It is notable that a normative text classifier trained on \GnG{} would do well on zero-shot transfer to the sci-fi corpus.
This makes it an attractive dataset for our experiments for the same reasons.

The {\em Plotto} dataset consists of 900~sentences extracted from a book, which catalogues plot points used in popular fiction. 
\citet{frazier2019learning} pruned some exceptionally anachronistic and misogynistic sentences from the corpora.
These sentences approximate the level of abstraction in the sci-fi corpus but have more contemporary narratives.

The {\em ROCstories}~\cite{rocstories} corpus contains 52,666 five-sentence stories, often about everyday life situations (e.g. going for a jog, taking a test in school, etc.). 
Unlike the previous two corpora, it covers a different space of more common, mundane events which usually do not have strong normativity connotations.

Sentiment is often used as a surrogate for normativity under the belief that non-normative behavior would be associated with negative sentiment.
The relationship between normativity and sentiment is not that simple, as we will show in  Section~\ref{sec:Experiment 3}.
We include sentiment experiments using large review datasets from IMDb, Yelp, and Amazon 
\cite{kotzias2015group}
because (1) previous value alignment research has incorporated sentiment analysis, and (2) we want to test our techniques on classifiers other than the normative text classifier.

Non-normativity is a superset of toxic language in the sense that toxic language is non-normative, but not all non-normative descriptions are toxic.
We also conduct experiments using toxic language classifiers - fine-tuned on sentiment corpora like the dataset from the Toxic Comment Classification Challenge\footnote{\url{https://www.kaggle.com/c/jigsaw\%2Dtoxic-comment-classification-challenge/}} as an alternative to the normative text classifier fine-tuned on \GnG{}.


\section{Normative Fine-Tuning}
\label{sec:Normative Fine-Tuning}

The GPT-2 model is trained by minimizing its cross-entropy loss given by \cite{radford2019language}:
\begin{align}
loss_{{\rm{w}}}(X, y) = &-\log\left(\frac{\exp(X[y])}{\sum_{i \in V} \exp(X[i])}\right) \nonumber\\
                    = &-\log\left(\sigma(X)_y\right)\label{eq:loss_word}
\end{align}
where $X$ is a vector containing output logits 
and \textit{y} is the index of the word from the ground truth in $X$.
$V$ is the model's vocabulary, $\sigma$ is the softmax function, and $\sigma(X)_y$ is the ground truth probability of the word.

To punish GPT-2 for producing 
non-normative text, we use a normative text classifier
to evaluate the model's performance and produce a reward value, which is applied to the loss and back-propagated through GPT-2. 
Given that the normativity of a sentence can only be determined by reading the entire sentence, the classifier must therefore produce a single numeric value per sentence.
%
Specifically,
we augment the cross-entropy loss computation with predictions from the pre-trained classifier. 
We define the sentence loss as:
\begin{equation}
loss_{{\rm{s}}}(s)=\frac{1}{n} \sum_{j \in s}{\rm{loss}}_{{\rm{w}}}(X_{j}, y_{j}) + u(s)\label{eq:loss_sentence}
\end{equation}
where $s$ is the continuation sentence generated by the neural language model,
$n=|s|-1$ is the number of the words in the continuation sentence,
$X_j$ is the $j^{\rm th}$ logit vector, and $y_j$ is the ground truth index for the $j^{\rm th}$ word.
$u(s)$ is a function of the output of the classifier converted into a {\em punishment} value;
a value of zero indicates no punishment, and higher positive values indicating increasingly non-normative sentences.
The $loss_{\rm w}$ counter-balances the reward and prevents the generated texts from descending into incoherent fragments.

\begin{figure}[htb]
\centering
\includegraphics[scale=0.6]{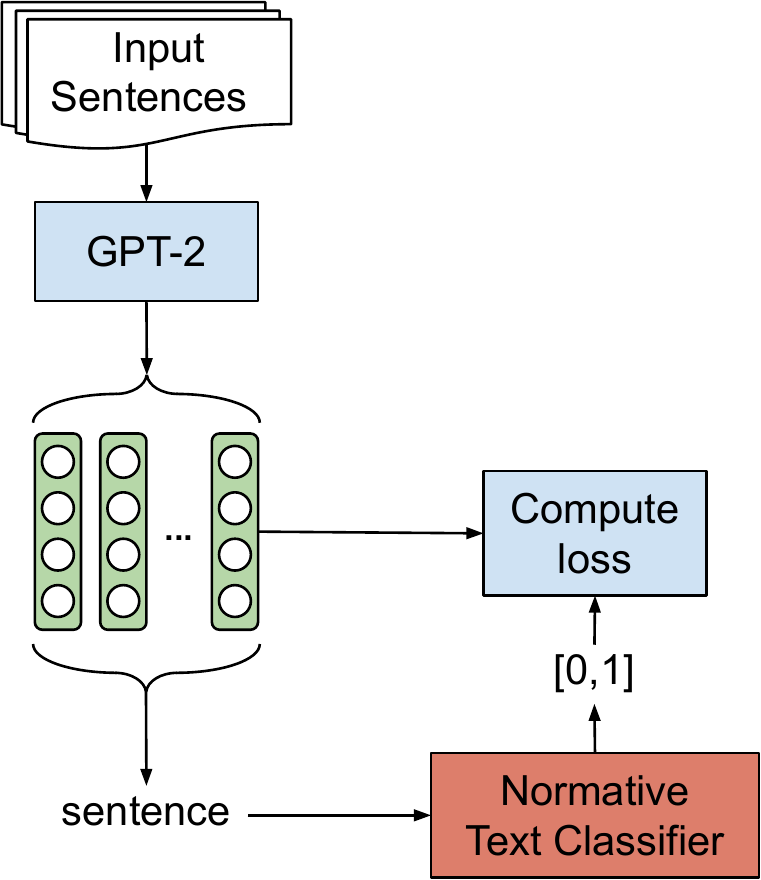}
\caption{Pipeline for fine-tuning GPT-2 with the classifier.
Loss is backpropagated through the output logits to GPT-2. 
}
\label{fig:pipeline}
\end{figure}

The punishment function $u(s)$ generates a value proportional to the average word loss so that it does not become overwhelmed by $loss_{\rm w}$:
\begin{equation}
u(s)=\rho\beta(1-C(s))(\frac{1}{n}\sum_{j\in s}{loss_{\rm w}(X_j,y_j))}
\label{eq:scifi_reward}
\end{equation}
where $s$ is a continuation sentence, 
$C(s)$ is the binary $\{0,1\}$ label given by the normative classifier, 
$\rho$ is a hyper-parameter to control the strength of the penalty, and 
$\beta=(1 - i \times 0.05)$ decreases the penalty as the number of fine-tuning iterations $i$ increases.
That is, if the generated sentence is classified as normative, a loss close to zero will be applied to each logit generated by the language model. 
If the generated sentence is non-normative, a higher total sentence loss will be applied to each logit.
$\beta$ decreases the step size during back-propagation to avoid over-shooting the local minima.
$loss_{\rm w}$ acts as a cycle loss component, punishing the sentences with undesirable characteristics. 

The fine-tuning process is as follows:
given a set of input sentences from a corpus, GPT-2 is used to generate successor sentences.
We generate 60 tokens and truncate at the first punctuation mark (e.g. periods, question marks).
These continuation sentences are fed through a classifier, which outputs the binary label we treat as a reward $C(s) \in \{0, 1\}$. Sentences labeled as $0$ are those with undesirable characteristic.
As per Equation~\eqref{eq:scifi_reward}, the reward is used to calculate the punishment score by subtracting from $1.0$ and scaling by the average word loss of the sentence.
This value is then used to compute a sentence loss as in Equation~\eqref{eq:loss_sentence}.
The sentence loss is averaged to obtain the token-level loss, which is then added to each logit from the continuation sentence and the loss is back-propagated into GPT-2.
The process is illustrated in Figure~\ref{fig:pipeline}.

To prevent the model from deviating too much from the language in the original dataset, we feed the fine-tuned model with the same set of input sentences at every loop and use the output sentences to even further fine-tune the model. As the model is trained, the output sentences will differ, and the 
reward value $C(s)$
may change after every iteration as the model shifts its distribution.


\section{Experiments}

We conduct three sets of experiments to (1)~verify the normative text classifier on generated continuations, (2)~evaluate our reward-based fine-tuning with the normative text classifier, (3)~evaluate our reward-based fine-tuning on other classifiers.

%

%


\subsection{Experiment 1: Replication of the Normative Classifier}
\label{sec:Experiment 1}


\begin{table}
\centering
\footnotesize
\begin{tabular}{lccc}
 & \textbf{Accuracy}& \textbf{Accuracy} & \textbf{\# of} \\ 
 \textbf{Dataset} & \textbf{(continuations)}& \textbf{(test corpora)} &\textbf{sent.} \\\hline
Plotto & 81.25 &89.67& 100\\
Sci-fi & 82.11 & 87.51& 300\\
\hline
ROCstories & 90.57 & 94.56& 100\\
Toxic & 86.84 &94.27& 400\\
Sentiment & 88.14 &93.90& 200\\
\hline
\end{tabular}
\caption{Results of Mechanical Turk study. Accuracy on generated continuations equals to the percentage of Mechanical Turk worker labels equivalent to labels produced by the normative classifier when classifying generated sentences, since Mechanical Turk worker labels are considered as ground truth label of generated continuations. Accuracy on original corpora is measured by the classifier on the held-out test sets of corpora sentences.}
\label{table:result_turk}
\end{table}

The normative text classifier by \citet{frazier2019learning} was evaluated on original sentences from a number of corpora, including the science fiction story corpus (sci-fi) we use in subsequent evaluation experiments.
Generated text potentially constitutes a shift in the text distribution. Hence, the accuracy of classifiers on generated continuations must be validated.

The 117M parameter GPT-2 is fine-tuned with training sets from Plotto, ROCstories, sci-fi, Toxic and Sentiment datasets, separately, in order to shift the output probability distribution of GPT-2 and to make it generate text similar to the corpus we used for training. 
%
The sci-fi and Plotto corpora were used for fine-tuning two different versions of the normative text classifier, starting with the classifier by~\citet{frazier2019learning}.
Thus the original classifier originally trained on \GnG{} was updated to the respective domains; a few-shot transfer paradigm.
We fine-tuned the classifiers for 2-5 iterations.

For the ROCstories, Toxic and Sentiment datasets, we directly train a BERT-based classifier on the given labels instead of fine-tuning the classifier that was first trained on \GnG.
We found the \GnG-trained classifier did not transfer well to ROCstories and thus collected our own normative and non-normative labels.
Toxic and Sentiment experiments do not look at normativity so we did not use the normative classifier.

A human participant study was then conducted to validate the fine-tuned classifier's accuracy.
Sentences from each corpus test set were randomly chosen and used as prompts for GPT-2 to generate continuation sentences.
70 crowd workers on Mechanical Turk labeled those generated sentences as normative or non-normative (or positive or negative sentiment, or toxic or non-toxic). 
Each sentence received at least three labels.
We treat the majority label from humans participants as the ground-truth.


%


Table~\ref{table:result_turk} shows the accuracy of classifiers on generated continuations and on sentences directly from the test sets. 
Accuracy decreases on generated continuations, but are on par with accuracy on sentences taken directly from the test corpora, and on par with the results from \citet{frazier2019learning}.
This indicates that any distributional shift during generation is likely inconsequential and the classifer achieves good zero-shot transfer to more datasets.

\subsection{Experiment 2: Decreasing Non-Normative Generation}
\label{sec:Experiment 2}



In this set of experiments, we seek to determine if, and by how much, the amount of non-normative behavior descriptions generated by GPT-2 decreases when fine-tuned with the normative text classifier.
We emphasis the decrease of non-normative language because both normative and neutral languages are acceptable.

Consistent with Experiment \#1, we first fine-tune the 117M parameter version of GPT-2 with sentences sampled from three datasets: ROCstories, sci-fi, and Plotto.
%
%
This gives us 
three versions of GPT-2: {\em GPT-ROCstories}, {\em GPT-scifi} and {\em GPT-plotto}, respectively.
The 117M GPT-2 model is fine-tuned for two, three and five iterations separately on ROCstories, Plotto and sci-fi to avoid overfitting.
%
%
We then fine-tune each of these models a second time using the reinforcement learning, reward-based technique in Section~\ref{sec:Normative Fine-Tuning}. 
We refer to these models as {\em GPT-ROCstories-norm}, {\em GPT-sci-fi-norm}, and {\em GPT-Plotto-norm}, respectively.
%
%
Due to the small size of the datasets, 
GPT-2 easily overfits during training. Therefore, we only fine-tune one of its 12 attention heads to avoid overfitting.


We evaluate the performance of our fine-tuned {\em GPT-$X$-norm} models ({\em $X$=plotto}, {\em scifi}, {\em ROCstories}) by analyzing the change in the proportion of generated text that is non-normative.
We measure the ratio of non-normative to normative text in two ways.
First, we use the normative text classifier on continuations generated by baselines and fine-tuned models.
This is an automated evaluation; Experiment 1 suggests the normative text classifier has high accuracy on the continuations.
However, the gold standard is the human participant labels.
For our second evaluation metric, 
we hired 70 Mechanical Turk workers to label generated continuation sentences as normative (including neutral) or non-normative.
At least 3 crowd workers labeled each sentence and the majority vote is considered as the ground truth label.

{\setlength{\tabcolsep}{0.45em}
\begin{table}[t!]
\centering
\footnotesize
\begin{tabular}{ll l l l}
 & \multicolumn{2}{c}{\bf \% non-norm.} &  &  \textbf{Test}\\
  \textbf{Model} & \textbf{Auto} & \textbf{Human} & \textbf{Perpl.} &\textbf{size}\\ \hline
GPT-ROCstories & 64.15&58.49 & 81.097 &50\\
GPT-ROCstories-norm & 26.42 &22.64 & 82.958 &50\\
\hline
GPT-Plotto & 81.25& 72.91& 34.271 &50\\
GPT-Plotto-norm & 59.18 & 53.06 & 32.322 &50\\
\hline
GPT-scifi & 35.11& 26.58 & 23.885 &300\\
GPT-scifi-norm & 15.79 & 18.27& 24.522 &300\\
\hline
\end{tabular}
\caption{The proportion of non-normative behavior and events (\% Non-norm) generated by different fine-tuned models on different datasets. 
Ratios are measured using the normative text classifier (automated) and Mechanical Turk studies (human labeling).
}
\label{table:result_normative}
\end{table}
}

Table~\ref{table:result_normative} shows the proportions of non-normative sentence continuations for both the baseline and the fine-tuned models. 
The results are summarized below.
%
%
We note percentage decreases, which are the relative percentage decrease compared to the original statistics.\footnote{Percentage decrease is calculated by $(p - \hat p) / p$, where $p$ and $\hat p$ are the proportion of non-normative behavior (\% Non-norm) generated by \textit{GPT-$X$} and \textit{GPT-$X$-norm}, respectively. }
\begin{itemize}
\item {\em GPT-ROCstories} 
generates non-normative continuations $64\%$ of the time according to the normative text classifier, which reduces to $26\%$ after further fine-tuning, a $59\%$ decrease.
Humans label {\em GPT-ROCstories} continuations as non-normative $58\%$ of the time, which drops to $22\%$, a $61\%$ decrease.
\item {\em GPT-Plotto} 
generates non-normative continuations $81\%$ of the time according to the normative text classifier, which reduces to $59\%$ after further fine-tuning, a $27\%$ decrease.
Humans label {\em GPT-Plotto} continuations as non-normative $72\%$ of the time, which drops to $53\%$, also a $27\%$ decrease.
\item {\em GPT-scifi}
generates non-normative continuations $35\%$ of the time according to the normative text classifier, which reduces to $15\%$ after further fine-tuning, a $55\%$ decrease.
Humans label {\em GPT-scifi} continuations as non-normative $26\%$ of the time, which drops to $18\%$, a $31\%$ decrease.
\end{itemize}
We observe that the classifier results are generally in line with the human evaluation results. 
The models fine-tuned on the Plotto dataset generally generate more non-normative continuation sentences and are more difficult to induce normativity. 
The models fine-tuned on the sci-fi dataset have the lowest frequency of non-normative generations, but can still be induced to produce lower frequencies with this method.

The perplexity remains steady after fine-tuning with the normative text classifier, indicating that the {\em GPT-$X$-norm} models are not overfitting nor losing their fluency.
Table~\ref{table:example} shows some examples of sentences generated by the fine-tuned GPT-2 models for the sci-fi and Plotto domains.

\begin{table}[t!]
\centering
\footnotesize
\begin{tabular}{lp{2.2in}}
\textbf{Label} & \textbf{Sentence}\\ \hline
Non-norm. & Mollari now refuses to pay the two parents' expenses and lives.\\
\hline
Non-norm. & Garibaldi slaps the door behind them and locks it behind them.\\
\hline
Non-norm. & He considers himself morally superior to his family because he is wealthy.\\
\hline
Norm. & Nathaniel repays his debt through an honest act of honest enterprise.\\
\hline
Norm. & He then makes a generous and appropriate sacrifice.\\
\hline
Norm. & He returns home to support his country.\\
\hline
\end{tabular}
\caption{Examples of generated normative and non-normative sentences from {\em GPT-Plotto} and {\em GPT-scifi}.}
\label{table:example}
\end{table}

\subsection{Experiment 3: Other Classifiers}
\label{sec:Experiment 3}


In the previous sections we evaluate how the normative text classifier and our fine-tuning technique work together to decrease the generation of non-normative text. In this section, we ablate our system and evaluate our fine-tuning method independently of the normative text classifier.
We seek to determine whether the fine-tuning technique is general enough to work with other classifiers.

We replicate the experimental methodolgy in  Section~\ref{sec:Experiment 2} but with the Toxic Comment Classification dataset and the Sentiment dataset. 
Sentiment is often used as a surrogate for normativity because non-normative behavior might be inferred to be perceived with negative sentiment, and because labeled sentiment data is more readily available.
Toxic language is a subclass of non-normative behavior.

First, we train two classifiers using the same technique as~\citet{frazier2019learning}.
Specifically, we fine-tune BERT on the datasets with ground-truth sentiment and toxicity labels. 
Both classifiers are fine-tuned 5 times on training set of corpora and tested on test sets (see Table~\ref{table:result_turk}). 

We follow the methodology in Section~\ref{sec:Experiment 2} to produce new GPT-2 baseline models.
%
To make the improvement from applying our technique more visible, we fine-tuned the 117M GPT-2 with only toxic or negative sentences when producing the baseline models, and obtained {\em GPT-senti-ext} and {\em GPT-toxic-ext}, which frequently produce negative or toxic generated continuations. We also fine-tuned the 117M GPT-2 with balanced datasets (half negative texts and half toxic texts, respectively), and refer to these two models as {\em GPT-senti-bal} and {\em GPT-toxic-bal}.
We then further fine-tune these models using their respective classifiers per the technique in Section~\ref{sec:Experiment 2}.

\begin{table}[t!]
\centering
\footnotesize
\begin{tabular}{lllll}
 & \multicolumn{2}{c}{\textbf{\% Non-norm}} & & \textbf{Test}  \\ 
\textbf{Model}&\textbf{Auto} & \textbf{Human} & \textbf{perpl.}& \textbf{size}  \\
\hline
GPT-toxic-ext & 59.03 &57.01 &54.819 &200\\
GPT-toxic-ext-norm & 27.75 &30.70 &62.302 &200  \\
\hline
GPT-senti-ext & 71.13&76.29 & 83.717 & 100  \\
GPT-senti-ext-norm & 45.36 &42.27&83.443 & 100  \\
\hline
GPT-toxic-bal & 37.89 & 33.04 & 50.535 & 200\\
GPT-toxic-bal-norm & 24.79 & 28.76 & 60.100 & 200\\
\hline
GPT-senti-bal & 44.33 & 36.08 & 91.927 & 100\\
GPT-senti-bal-norm & 35.05 & 29.90 & 90.261 & 100\\
\hline
\end{tabular}
\caption{The proportion of non-normative behavior and events (\% non-norm.) generated by different fine-tuned models on different datasets. 
}
\label{table:result_exp_3}
\end{table}

Table~\ref{table:result_exp_3} shows the percentage of textual continuations that are either toxic or contain negative sentiment. 
Fine-tuning {\em GPT-senti-ext} with the sentiment classifier can reduce negative sentiment from $76\%$ to $42\%$, a $45\%$ reduction. 
Fine-tuning {\em GPT-senti-bal} with the sentiment classifier can reduce negative sentiment from $36\%$ to $29\%$, a $17\%$ reduction.
Fine-tuning {\em GPT-toxic-ext} with the toxic classifier can reduce toxic language from $57\%$ to $30\%$, a $46\%$ reduction.
Fine-tuning {\em GPT-toxic-bal} with the toxic classifier can reduce toxic language from $33\%$ to $28\%$, a $13\%$ reduction.
This shows that the fine-tuning technique working on sentence loss is agnostic to which classifier is used.


We did not compare our results directly to the Plug and Play Langauge Models (PPLM) work~\cite{dathathri2019plug}, which also uses a toxic word classifier to fine-tune a language model.
PPLM fine-tunes on a word-by-word basis instead of at the sentence unit, making it difficult to account for the context needed for determining normativity.
For toxic language reduction, their model is trained to operate on a pre-given set of prompts such as ``black'' or ``asian''. For these prompts, given during training, they can reduce GPT-2's toxic language frequency from $\sim$10\% to $\sim$6\%. 
This is non-significant ($p < 0.23$) though it is challenging to reduce a number that is already close to zero.
When we prompt our {\em GPT-toxic-bal} and {\em GPT-toxic-bal-norm} with ``asian'' and ``black'', we see reductions from $52\%$ to $32\%$ and from $70\%$ to $50\%$, respectively.
Our classifier has only ever seen the word ``black'' once and has never seen the word ``asian''.
We see higher occurrences of toxicity because GPT-2 is fine-tuned on the equal numbers of toxic and non-toxic sentences from the Toxic dataset (whereas PPLM compares against a non-fine-tuned version of GPT-2) and because GPT-2 has a pre-existing unjust bias toward these words. 



\begin{figure}[t!]
\centering
\centerline{\includegraphics[width=\columnwidth]{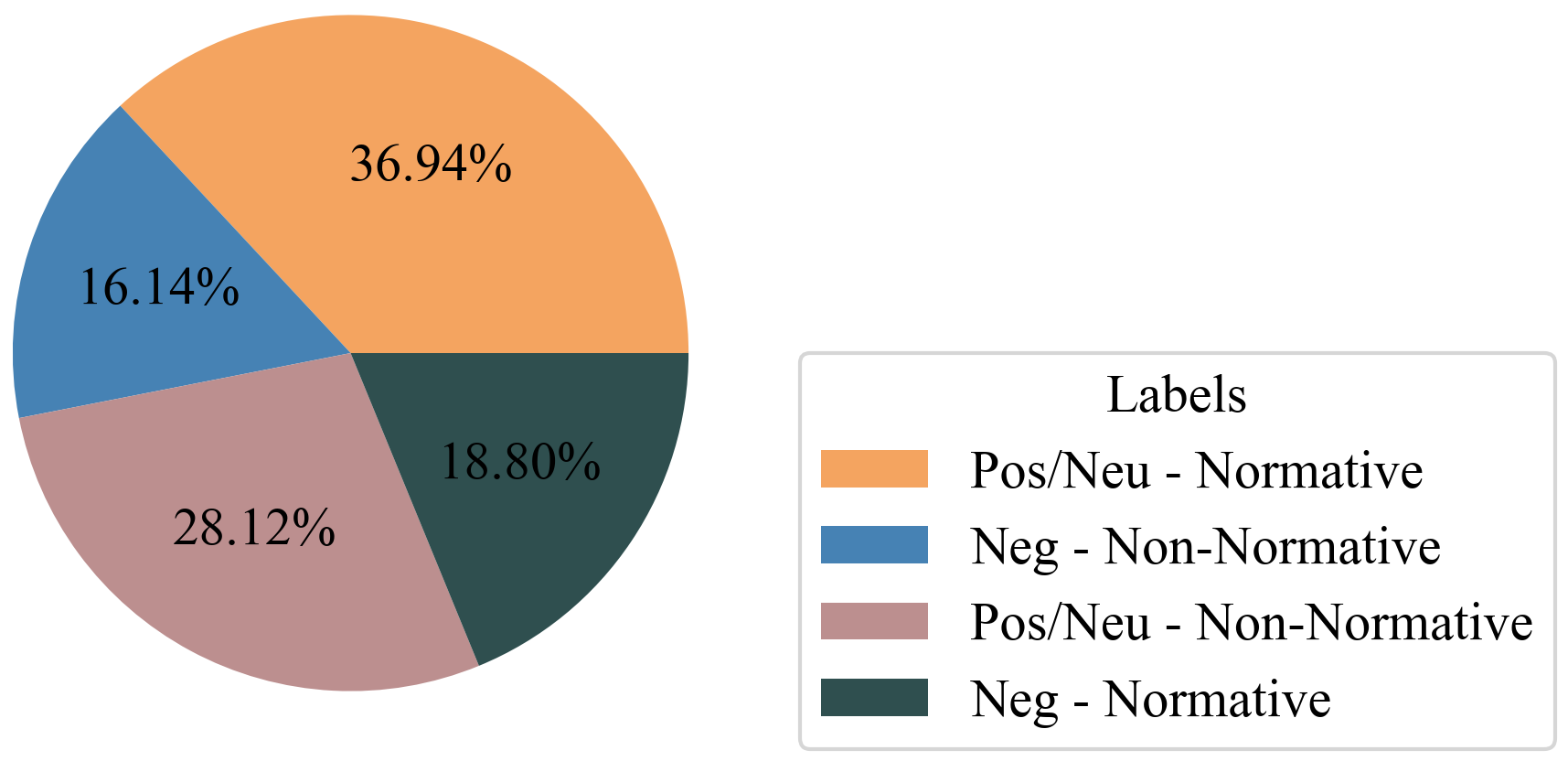}}
\caption{Differences between normative classification and sentiment classification.}
\label{fig:pie}
\end{figure}

To address the relationship between sentiment and normativity, we sample 300 sentences from the sci-fi corpus and 300 sentences from the generation results of the trained GPT-2 model and classify them using the normative text classifier and SentiWordNet~\cite{esuli2006sentiwordnet}.
Figure~\ref{fig:pie} shows the percentage of sentences were classified as 
(a)~both normative and positive/neutral sentiment (orange),
(b)~both non-normative and negative sentiment (blue),
(c)~normative but negative sentiment (green), and
(d)~non-normative but positive/neutral sentiment (brown).
Only about half the sentences tested (53.08\%) had sentiment and normativity labels that matched, whereas 46.92\% of sentences have conflicting labels.

\section{Discussion}



We demonstrate how value-aligned priors such as normative text classifiers can act as a reward provider to nudge the GPT-2 language model towards producing more normative and neutral descriptions of behaviors and events. Applying this reward-based fine-tuning technique reduces the likelihood of generating sentences containing non-normative behavior by approximately $\sim$27-61\%, depending on the dataset. 
Some datasets are more non-normative and more resistant to reduction of non-normativity. 
Datasets with low non-normativity to begin with are, naturally, more resistant.

Beyond the numerical results, this shows evidence that policy-gradient based reinforcement learning approaches to fine-tuning can be an effective means for reducing the generation of non-normative descriptions. 
%
By using a normative text classifier---a {\em value aligned prior}---one can fine-tune a language model to the desired domain and then fine-tune using the prior again to reduce non-normative generation that stems either from GPT-2 or from the domain corpus.
We provide this approach as an alternative---or in complement to---debiasing techniques that attempt to correct for prejudicial bias in datasets prior to training.
Our approach is roughly equivalent to teaching a language model to censor itself.

The policy-gradient based reinforcement learning technique using a value-aligned prior can be even more valuable with {\em GPT-3}, where fine-tuning to a domain is less necessary. 
GPT-3 has been demonstrated to be capable of non-normative, toxic, and prejudicially biased generation.

By using the normative text classifier trained on \GG{} comics, our results are limited to Western---and in particular American---mainstream ideals of normative behavior. 
We acknowledge that culture is not monolithic, even within the United States of America, and this represents only one of many possible sources of normativity.
We cannot conclusively say that our results will hold if we had normative text classifiers trained from different source materials.
Because general sources of normative behavior are currently hard to come by, we attempt to show generalization of our technique with experiments using sentiment and toxic language.




One limitation of our work is that the normative text classifier can only classify individual sentences without context. 
Given the context-dependent nature of normativity, the normative text classifier may overlook non-normative sentences that may appear normative out-of-context. 
This may lead to GPT-2 still producing this sentence in its non-normative context. 
Another limitation is that fine-tuning GPT-2 on Plotto, ROCstories and sci-fi datasets leads to it generating both neutral and normative sentences. If a model that generates solely normative sentences is desired, one can substitute the normative classifier with a ternary classifier with labels for normative, non-normative, and neutral sentences, and adjust the reward signals accordingly.
Furthermore, fine-tuning classifiers requires datasets with labeled exemplars, hence, in order to replicate our work to generate texts with some other desirable characteristics, datasets with labels would be prerequisites. 

The motivation of our work is to show how those who are concerned with generating undesirable text language models can obtain some control over the generation process.
We look at normativity, but also show how other criteria can be applied.
However, as is true for most algorithms, those with malicious intent can find ways to corrupt the intentions of the work. 
For example, equation~(\ref{eq:scifi_reward}) can be trivially modified to punish normative text instead of non-normative text.


\section{Conclusions}

We have shown that large-scale transformer-based neural language models can be made to generate text containing fewer descriptions of non-normative behavior 
by applying data-efficient, policy-gradient reinforcement learning.
As most large-scale language models are trained on datasets from the internet and from books, the potential for intentional or unintentional non-normative language persists.
We see this as a first step toward decreasing the potential for unintended, unacceptable, anachronistic or harmful language.

While our primary result is to show that we can decrease the generation of non-normative behavior descriptions, our normative classifier of choice is rooted in Western/American norms and values.
Normative classifiers are rare and datasets containing normative or preference learning examples are difficult to obtain.
However, our results show that even small datasets of normative examples can be converted into few-shot classifiers and applied to new domains.
By replicating our results with sentiment and toxic classifier, we show that our technique is not specific to any one classifier.


\bibliographystyle{acl_natbib}
\bibliography{acl2020}

\end{document}